# AI Matrix – Synthetic Benchmarks for DNN

Wei Wei, Lingjie Xu, Lingling Jin, Wei Zhang
Alibaba Inc
{w.wei, lingjie.xu, l.jin, wz.ww}@alibaba-inc.com

Tianjun Zhang
University of California, Berkeley
tianjunz@berkeley.edu

## INTRODUCTION

Deep neural network (DNN) architectures, such as convolutional neural networks (CNN), involve heavy computation and require hardware, such as CPU, GPU, and AI accelerators, to provide the massive computing power. With the many varieties of AI hardware prevailing on the market, it is often hard to decide which one is the best to use. Thus, benchmarking AI hardware effectively becomes important and is of great help to select and optimize AI hardware.

Unfortunately, there are few AI benchmarks available in both academia and industry. Examples are BenchNN[1], DeepBench[2], and Dawn Bench[3], which are usually a collection of typical real DNN applications. While these benchmarks provide performance comparison across different AI hardware, they suffer from a number of drawbacks. First, they cannot adapt to the emerging changes of DNN algorithms and are fixed once selected. Second, they contain tens to hundreds of applications and take very long time to finish running. Third, they are mainly selected from open sources, which are restricted by copyright and are not representable to proprietary applications.

In this work, a synthetic benchmarks framework is firstly proposed to address the above drawbacks of AI benchmarks. Instead of pre-selecting a set of open-sourced benchmarks and running all of them, the synthetic approach generates only a one or few benchmarks that best represent a broad range of applications using profiled workload characteristics data of these applications. Thus, it can adapt to emerging changes of new DNN algorithms by re-profiling new applications and updating itself, greatly reduce benchmark count and running time, and strongly represent DNN applications of interests. The generated benchmarks are called AI Matrix, serving as a performance benchmarks matching the statistical workload characteristics of a combination of applications of interests.

## 1. METHODOLOGY
### 1.1 General framework

This framework consists of 3 key steps shown below. For example, in CNN, the convolutional layers typically only consume around 5% of the memory, but contribute to around 90%-95% of the total computational workload at inference time [6]. We consider convolution only in our framework for simplicity.

**Application monitoring system**: a platform backend program. It is used to collect the neural network (NN) applications execution data with minimum overhead. It collects the data of every layer/operation executed with detailed parameters, e.g., number of convolutional execution, input size, input channel, kernel size, kernel stride.

**Workload analysis**: it deploys the nearest neighbor algorithm to cluster the operations according to each layer's parameters. The convolutional operations are clustered twice by input size and convolution filter size parameters as shown in Tab. 1.

**Workload synthesizer:** it is the most important step in our framework. We assembly the network for each group and connect each group to form a final synthetic network model. For each group, we generate the synthetic network with genetic algorithm [4] using fitness function defined in section 2.1. We set up this rule to define the networks and the number of each convolution filter can be encoded into fixed-length binary strings and optimized in genetic algorithm.

A synthetic DNN is composed of $S$ groups. And each group contains $Ts$ nodes where $s=0,1…S$. Each node represents one convolutional operation. The nodes are connected one by one as filter size decreases except that the output size is different (shown in group 1 of Fig.1). If the node (the first one of a group) has several inputs (of the same size), a concat operation (CA) is added to combine these inputs. After convolution, batch normalization (BN) and ReLU are followed [5]. We do not encode the fully connected part of a network as it becomes less important in the current models.

After each group is assembled, neighboring groups are connected via a spatial pooling (PL) operation, which will change the spatial size to match the input size of next group. The example below shows the generated synthetic DNN from section 3.1 of 3 classic CNN models. The last output is connected with fully connected (FC) and softmax layer.

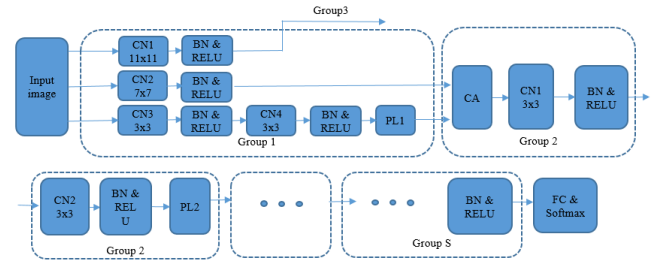

Fig.1 synthetic CNN using statistical data from Tab.1

### 1.2 Workload characteristics and fitness function

The NN architectures are usually complex with various components like blocks, layers and operators combined with different parameters. To extract meaningful workload characteristics from NN applications is an open topic and basis for our synthetic benchmarks. As we know, 90-95% of computation comes from convolutional operation and multiply-accumulate (MAC) operation is the only operation inside. The size of the input and filter size affect the final MACs of a convolutional operation. Therefore, we propose to consider following workload characteristics 1) statistical distribution of convolutional layer parameters 2) computational workload (MAC) 3) launched warps (WP) in GPU. We match 1) in synthetic benchmarks and apply 2) and 3) in fitness function.

$$Fitness = 0.5 * \frac{|MAC - MACreal|}{MACreal} + 0.5 * \frac{|WP - WPreal|}{WPreal}$$

As a result, generated CNN has similar statistical distribution of layer parameters, computational workload and launched warps of models of interests. The synthetic CNN model will represent the models of interests.

## 2. EXPERIMENT AND EVALUATION

### 2.1 Experiment on data from mixing of 3 classic models

As we know, almost all of the impactful models for CNN are originating from ILSVRC competition. We select 3 different classic models Alexnet, Vgg16, GoogleNet, running once for each model on our monitoring system and log the total inference execution data. After workload analysis, the layers' statistical information is collected and clustered as shown in Tab.1 below.

| Group | Group center (HxW) | Counts | filter size, filter stride |
|---|---|---|---|
| 1 | 224x224 | 1 | 11,4 |
|   |         | 1 | 7,2 |
|   |         | 2 | 3,1 |
| 2 | 112x112 | 2 | 3,1 |
| 3 | 56x56   | 4 | 3,1 |
|   |         | 1 | 1,1 |
| 4 | 28x28   | 4 | 5,1 |
|   |         | 5 | 3,1 |
|   |         | 8 | 1,1 |
| 5 | 14x14   | 3 | 5,1 |
|   |         | 13 | 3,1 |
|   |         | 22 | 1,1 |
| 6 | 7x7     | 2 | 3,1 |
|   |         | 10 | 1,1 |

Tab. 1. Statistical data of convolutional operations

Then, we deploy the genetic algorithm on each group by using the fitness function we set in section 2.2. After generating the synthetic model, we summarize the target fitness values of 3 real models and our synthetic benchmarks. As shown in Tab 2, the error is very small with 1.19% as maximum.

| Group | 3 models (MACs, warps) | Synthetic model (MACs, warps) | Fitness Error (%) |
|---|---|---|---|
| 1 | 2156022912,2802996 | 2156050176,2803928 | 0.02 |
| 2 | 2774532096,2792888 | 2774419200,2859720 | 1.19 |
| 3 | 4983881728,3184980 | 4986729216,3184876 | 0.03 |
| 4 | 5280485376,2221154 | 5280615536,2221030 | 0.00 |
| 5 | 2199505920,2789028 | 2199066688,2788961 | 0.01 |
| 6 | 109734912,885229 | 109735598,885337 | 0.00 |

Tab. 2. MACs and launched warps of each group

### 2.2 Experiment on data from Alibaba

The data comes from our AI platform that logs statistical data of different CNN models running on it. The synthetic benchmark could be generated based on data collected in any period of time that in turn represents the models running in that period of time. We collected 608 convolutional operations with input size ranging from 224x224 to 5x5. The filter size ranges from 6 to 1 with 3x3 and 5x5 are the most heavily used on our platform as shown in Fig. 2

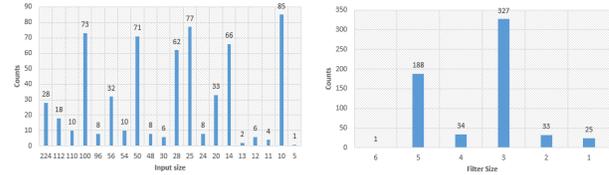

Fig. 2 Statistical data of convolutional operations

Similar to the experiment in section 3.1. We shrink the size of synthetic models proportionally and apply workload analyzer to cluster the data into 6 groups. After that, genetic algorithm is applied on each group to generate a synthetic model. As shown in Tab. 3, the errors between real models and the synthetic model range from 0.00-0.96% which is quite good.

| Group | Real models (MACs, warps) | Synthetic model (MACs, warps) | Fitness Error (%) |
|---|---|---|---|
| 1 | 3728928152,7139725 | 3729030144,7277052 | 0.96 |
| 2 | 6129559271,11169027 | 6129236736,11165590 | 0.01 |
| 3 | 14495976374,9817393 | 14500048640,9813450 | 0.03 |
| 4 | 11699424510,4706169 | 11699536710,4708200 | 0.02 |
| 5 | 3330293660,3374419 | 3330837398,3374523 | 0.00 |
| 6 | 258530570,575312 | 258472240,573424 | 0.17 |

Tab. 3. MACs and launched warps of each group

## 3. CONCLUSION AND FUTURE WORK

This work proposes a synthetic benchmark framework that better adapts to the emerging changes of DNN algorithms, has significantly less number of tests and running time, and is representable of proprietary applications. The framework is validated using mixing of 3 classic models and real data collected from Alibaba AI platform. The generated synthetic benchmark represents statistical distribution of layer parameters and workload characteristics of all the real applications of interests. Future work is to consider integrating the block architecture, e.g., inception module and residue module, into the synthetic models. RNN based models will also be considered for synthetic models.